\documentclass[11pt]{article}

\usepackage[final]{acl}

\usepackage{times}
\usepackage{latexsym}
\usepackage{algorithm}
\usepackage{algpseudocode}
\usepackage{amsmath}
\usepackage{amssymb} 

\usepackage{makecell}   

\usepackage[T1]{fontenc}

\usepackage[utf8]{inputenc}

\usepackage{microtype}

\usepackage{inconsolata}

\usepackage{graphicx}

\usepackage[table]{xcolor}
\usepackage{tabularx}
\usepackage{bm}
\usepackage{booktabs}
\usepackage{color}
\usepackage{multirow}
\usepackage{hyperref}

\title{Retrieving to Recover: Towards Incomplete Audio-Visual Question Answering via Semantic-consistent Purification
}

 \author{Jiayu Zhang$^1$\thanks{These authors contributed equally.} , Shuo Ye$^{1,3}$\textsuperscript{*}, Qilang Ye$^{4}$, Zihan Song$^5$, Jiajian Huang$^1$, Zitong Yu$^{1,2}$\thanks{Corresponding Author}\\
         $^{1}$School of Computing and Information Technology, Great Bay University  \\ $^{2}$Dongguan Key Laboratory for Intelligence and Information Technology \\ $^3$Tsinghua Shenzhen International Graduate School, Tsinghua University \\$^4$College of Computer Science, Nankai University \\ $^5$School of mathematics, Sun Yat-sen University\\ qmmcxm1019@gmail.com, yuzitong@gbu.edu.cn}

\begin{document}

\maketitle
\begin{abstract}
Recent Audio-Visual Question Answering (AVQA) methods have advanced significantly.  However, most AVQA methods lack effective mechanisms for handling missing modalities, suffering from severe performance degradation in real-world scenarios with data interruptions. Furthermore, prevailing methods for handling missing modalities predominantly rely on generative imputation to synthesize missing features. While partially effective, these methods tend to capture inter-modal commonalities but struggle to acquire unique, modality-specific knowledge within the missing data, leading to hallucinations and compromised reasoning accuracy. To tackle these challenges, we propose R$^2$ScP, a novel framework that shifts the paradigm of missing modality handling from traditional generative imputation to retrieval-based recovery. Specifically, we leverage cross-modal retrieval via unified semantic embeddings to acquire missing domain-specific knowledge. To maximize semantic restoration, we introduce a context-aware adaptive purification mechanism that eliminates latent semantic noise within the retrieved data. Additionally, we employ a two-stage training strategy to explicitly model the semantic relationships between knowledge from different sources. Extensive experiments demonstrate that R$^2$ScP significantly improves AVQA and enhances robustness in modal-incomplete scenarios. \footnote{Our code is available at \href{https://github.com/AoKoo233/Incomplete_AVQA}{[GitHub Repo]}}

\end{abstract}

\section{Introduction}

In the rapidly evolving landscape of multimodal understanding, Audio-Visual Question Answering (AVQA)~\cite{zhao2025audio,pei2025guiding} has emerged as a pivotal task, requiring models to reason across visual, audio, and textual domains to achieve a comprehensive understanding of dynamic scenes. By effectively synthesizing heterogeneous information, AVQA systems have demonstrated remarkable potential in applications ranging from intelligent assistants to video content analysis. Despite these advancements, achieving robustness in AVQA models for real-world deployment remains a significant challenge~\cite{wu2024deep}. While standard methods typically assume modal completeness, practical scenarios often violate this assumption due to issues like device malfunctions, sensor occlusion, or data transmission failures. In cases where a critical modality is unavailable, such as the loss of audio in a musical performance video, the performance of conventional models tends to deteriorate significantly.

\begin{figure}[!t]
  \centering
  \includegraphics[width=\linewidth]{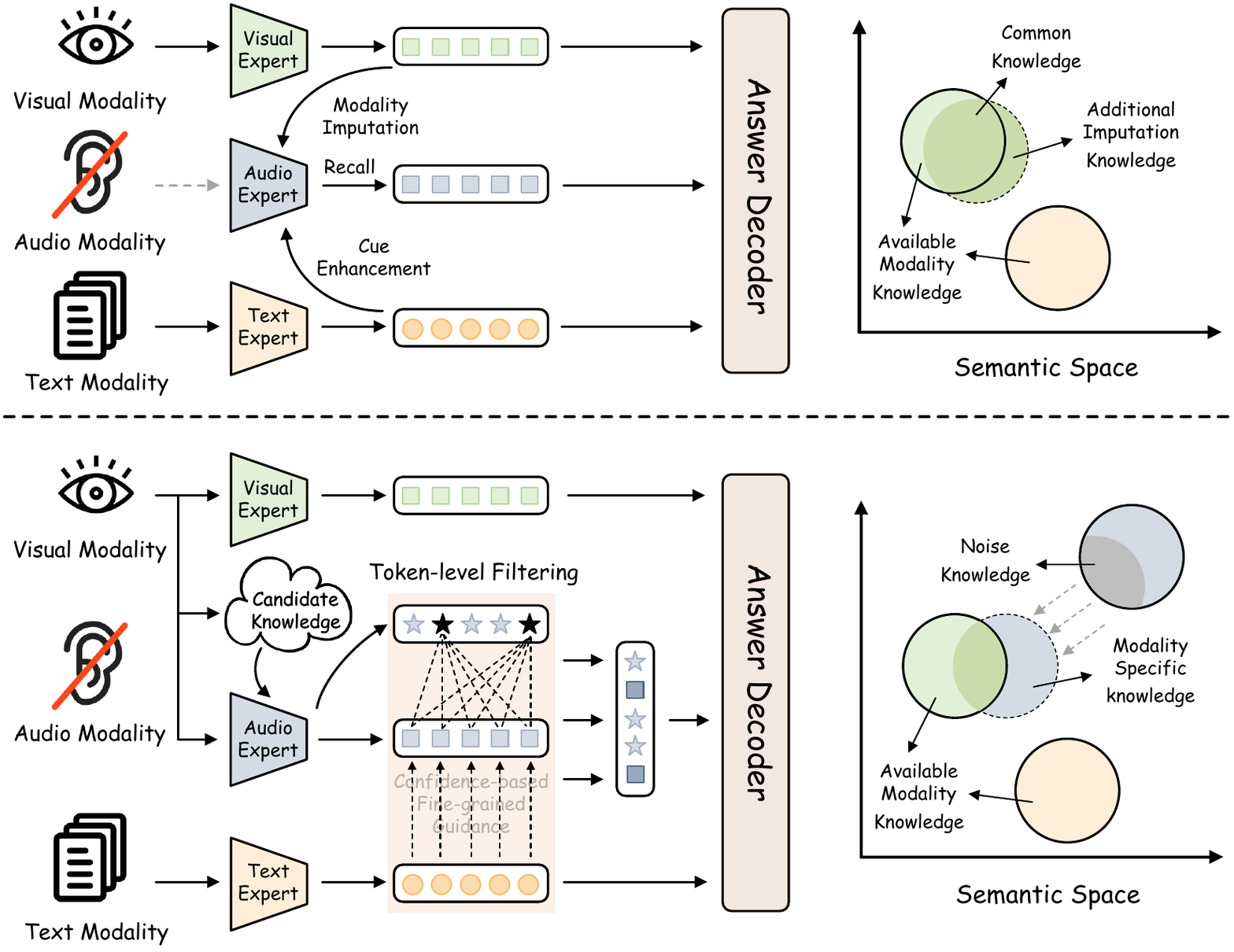}
  \vspace{-1.5em}
  \caption{Comparison of traditional methods~(top) and R$^2$ScP~(bottom). Traditional methods typically rely on modality imputation but often yield redundant common knowledge that lacks distinctiveness. In contrast, R$^2$ScP preserves valuable modality-specific knowledge from the candidate pool while effectively suppressing the inherent noise knowledge.}
  \label{fig:1}
\vspace{-1em}
\end{figure}

To mitigate the impact of incomplete data, the prevailing research~\cite{chen2025i3,zhu2025bridging,xin2025i2moe,zhang2025synergistic} paradigm has largely focused on generative modality imputation. Pioneering works, such as the relation-aware missing modal generator proposed by Park et al.~\cite{park2024learning}, attempt to synthesize pseudo-features for the missing modality by conditioning on the available data. While these generative methods have shown promise, they face an intrinsic limitation regarding semantic hallucination and noise. As illustrated in Figure~\ref{fig:1}~(top), generative models tend to produce common knowledge features, which are generic representations that lack the fine-grained, modality-specific details required to answer complex questions. For example, when inferring missing audio from a visual scene of a concert, a generative model might synthesize a generic ``music'' embedding but fail to capture the distinct timbre of the specific instruments visible, thereby introducing semantic noise that confuses the reasoning process.

In this paper, we challenge the dominance of generative imputation and propose a paradigm shift from generation to retrieval. We hypothesize that instead of synthesizing imperfect hallucinations, it is more effective to recall high-quality, real-world feature segments from a semantic database that are coherent with the available context. To this end, we present R$^2$ScP~(\textbf{R}etrieving to \textbf{R}ecover via \textbf{S}emantic-\textbf{c}onsistent \textbf{P}urification), a novel framework designed to achieve robust AVQA performance under missing modality conditions.

In contrast to current approaches, R$^2$ScP leverages a unified semantic space to retrieve candidate features for the missing modality based on the available inputs. However, raw retrieval inevitably introduces irrelevant information. To address this, we introduce a Context-aware Adaptive Purification mechanism~(CAP). CAP acts as a semantic filter, and it utilizes the semantic consistency between the retrieved candidates and the available modalities to verify and purify the retrieved features. By adaptively suppressing noise and highlighting contextually relevant cues, CAP ensures that only the information strictly beneficial for the QA task is integrated. Furthermore, we employ a mixture of experts~(MoE) strategy using a two-stage training process of independent expert training followed by expert mixing to explicitly model the complex inter-dependencies between retrieved knowledge and original inputs. Our main contributions are summarized as follows:
\begin{itemize}
    \item We propose R$^2$ScP, a novel framework that shifts the perspective of missing modality handling in AVQA from generative imputation to retrieval-based recovery, effectively preserving modality-specific details.
    \item We introduce the Context-aware Adaptive Purification mechanism~(CAP), which dynamically filters semantic noise from retrieved features by enforcing consistency with available modalities, ensuring high-fidelity feature reconstruction.
    \item Extensive experiments on multiple AVQA datasets demonstrate that our method significantly outperforms state-of-the-art competitors, achieving superior robustness in diverse missing modality scenarios.
\end{itemize}

\section{Related Work}
\subsection{Audio-Visual Question Answering}
Audio-visual question answering~(AVQA)~\cite{ye2026eyes} is a multimodal reasoning task that requires aligning visual, acoustic, and textual information for comprehensive scene understanding. Benefiting from advances in deep learning~\cite{zhang2026multimodal,wang2026micro,ye2026sugar,zhang2025mgtr,tang2025srvc,lin2025svc,xie2024fusionmamba,ye2026ika2,zhu2026H-GAR,zhu2026deltavlapriorguidedvisionlanguageactionmodels,zhu2025emosym,zhu2025uniemo}, recent AVQA methods improve multimodal understanding through spatiotemporal modeling. PSTP-Net~\cite{li2023progressive} progressively selects key spatiotemporal regions to localize question-relevant segments. QA-TIGER~\cite{kim2025question} employs a gaussian-based mixture-of-experts framework to capture continuous temporal dependencies and inject question context during encoding. AV-Master~\cite{zhang2025av} further adopts a dual-path design with dynamic adaptive focus sampling and global preference activation to reduce redundant audio-visual information.

Despite these advances, most AVQA methods assume complete modalities and rely heavily on audio-visual interaction, causing severe performance degradation in real-world scenarios with sensor malfunction or transmission failure. Unlike previous works, this paper focuses on handling the problem of missing modalities so that the model can perform robust inference even when key modalities are absent.

\subsection{Incomplete Multimodal Learning}
Incomplete multimodal learning aims to learn robust representations from partial observations in the presence of sensor failure or data corruption. Existing generative imputation methods mainly include reconstruction-based and representation-based generation. Reconstruction-based methods synthesize missing data from available modalities~\cite{tran2017missing,cai2018deep}. Early approaches typically use GANs or autoencoders to reconstruct raw data or feature maps. For example, MMIN~\cite{zhao2021missing} uses cascaded residual autoencoders to predict missing features from cross-modal associations. With diffusion models, IMDer~\cite{wang2023incomplete} recovers missing emotion cues via score-based generation. IMOL~\cite{zeng2025imol} further introduces cognitive memory replay, using cross-modal consistency to imagine missing modalities and retrieval-augmented contrastive learning to improve domain generalization.

Another line of work learns modality-invariant or disentangled representations for robustness under missing inputs. ShaSpec~\cite{wang2023multi} disentangles modality-shared and modality-specific features, while transformer-based models such as mmFormer~\cite{zhang2022mmformer} and ViLT~\cite{ma2022multimodal} use attention masking to fuse available tokens. Mixture-of-experts methods also offer flexibility: MoMKE~\cite{xu2024leveraging} preserves modality-specific knowledge with unimodal experts, and SimMLM~\cite{li2025simmlm} adopts dynamic gating with a ``More vs. Fewer'' ranking loss to handle varying modality availability.

Despite these promising advancements, current methods still face intrinsic limitations regarding semantic fidelity. Generative approaches often suffer from ``hallucination'', producing generic, common-knowledge features that severely lack the fine-grained, instance-specific details required for complex reasoning (e.g., generating a generic instrument sound instead of a specific violin timbre). While IMOL incorporates retrieval, it primarily uses it for contrastive alignment rather than explicit direct feature recovery. In contrast to these generative paradigms, our work proposes a paradigm shift towards retrieval-based recovery. By retrieving high-quality, real-world feature segments from a unified semantic space and applying context-aware purification, R$^2$ScP effectively mitigates semantic noise and preserves the distinctiveness of the missing modality, ensuring robust reasoning in AVQA tasks.

\section{Methodology}

\subsection{Problem Definition}

\begin{figure*}[!t]
  \centering
  \includegraphics[width=0.95\linewidth]{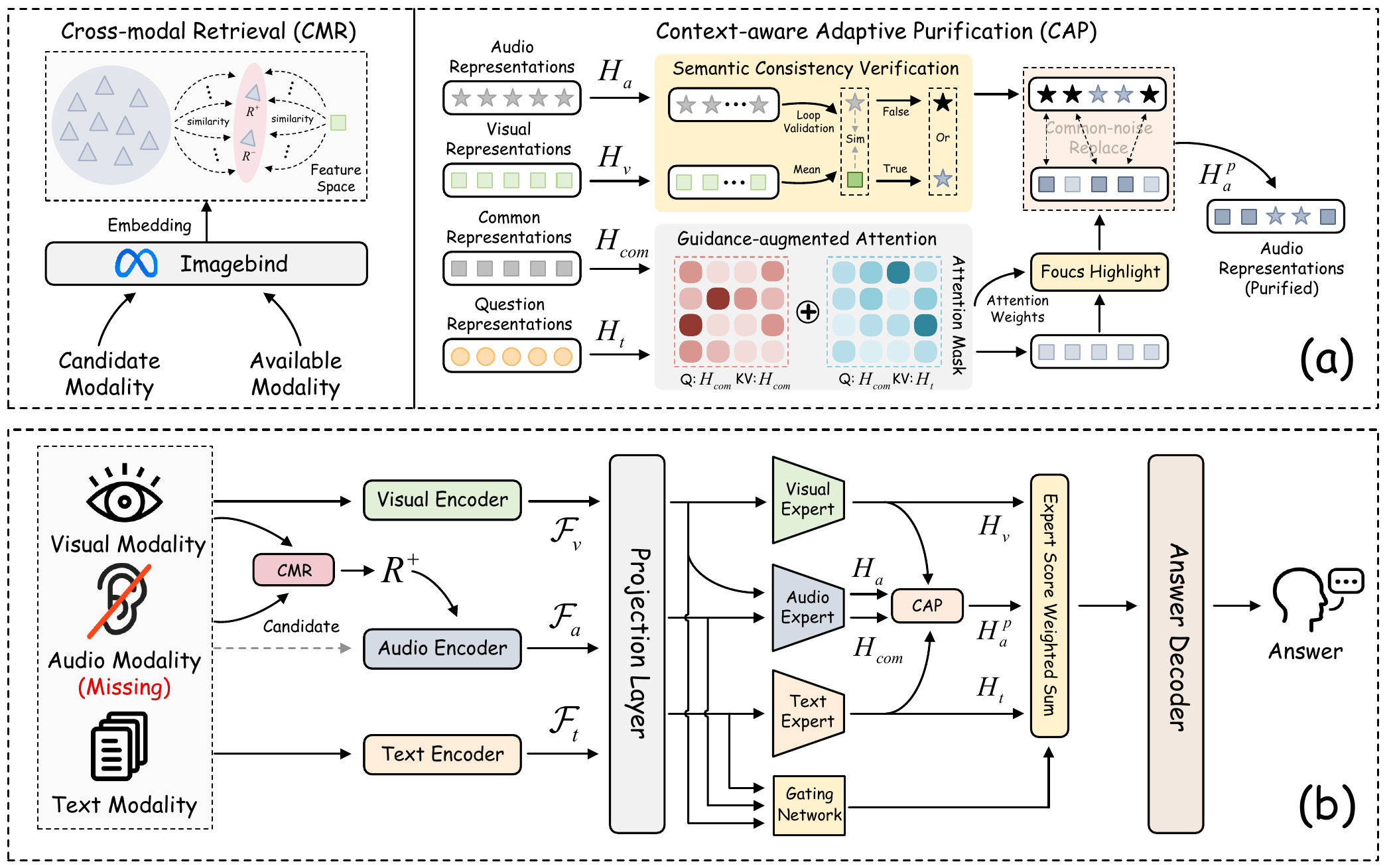}
  \caption{Overview of the proposed R$^2$ScP framework~(when the audio modality is missing). (a)~The CMR module retrieves candidate features from a unified semantic space, while the CAP mechanism acts as a semantic filter that refines the coarse retrieved features using the common knowledge between the visual and audio modalities. (b)~The overall architecture processes available and purified representations for the answer decoder.}
  \label{fig:2}
  \vspace{-1em}
\end{figure*}

The AVQA task aims to infer an answer $y \in \mathcal{Y}$ given a multimodal input sequence consisting of visual frames $V$, audio segments $A$, and a textual question $T$. In real-world scenarios, we encounter an incomplete modality setting where a subset of modalities may be missing or corrupted. Let $M = \{v, a, t\}$ denote the set of all modalities, and $M_{avl} \subseteq M$ denote the set of available modalities. The input features can be represented as $F = \{f_m | m \in M_{avl}\}$, where $f_m$ denotes the feature representation of modality $m$. Unlike previous works that rely on generative models $G(f_{avail}) \rightarrow \hat{f}_{miss} \rightarrow y$ to hallucinate missing features, our goal is to retrieve and purify real-world semantic knowledge. We propose R$^2$ScP, which learns a mapping $\Phi(M_{avl}) \rightarrow y$ by retrieving missing evidence from a unified semantic space and filtering it via a context-aware mechanism. The overall architecture of R$^2$ScP is shown in Figure~\ref{fig:2}.

\subsection{Cross-Modal Retrieval~(CMR)}

To bridge the semantic gap caused by missing modalities, we propose a retrieval-based recovery paradigm. We construct an external memory bank $\mathcal{B} = \{(\mathbf{k}_i, \mathbf{v}_i)\}_{i=1}^{M}$ using a unified semantic space (generated by a pre-trained multimodal model, e.g., Imagebind~\cite{girdhar2023imagebind}), where $\mathbf{k}_i$ represents the key embedding of a potential missing modality sample (e.g., audio) and $\mathbf{v}_i$ is its corresponding raw feature representation. Given an input with a missing modality (e.g., missing audio), we utilize the available modality (e.g., visual) as the query $\mathbf{Q}_{avl}$. We measure the semantic similarity between the query and the memory bank keys using the cosine similarity metric:
\begin{equation}
    \small
    \begin{aligned}
        S_{i} = \frac{\mathbf{Q}_{avl} \cdot \mathbf{k}_i}{\|\mathbf{Q}_{avl}\| \|\mathbf{k}_i\| + \epsilon} \\
    \end{aligned}
\end{equation}
we then subsequently retrieve the top-$n$ candidate set $\mathcal{R} = \{\mathbf{r}_i\}_{i=1}^{n}$ corresponding to the indices with the highest similarity scores $S$. These selected candidates serve as the raw semantic supplement for the missing modality.

\vspace{-0.3em}
\subsection{Context-aware Adaptive Purification}

Although the retrieved candidate set $\mathcal{R}$ provides potential semantic prototypes for the missing modality, raw retrieval inevitably introduces semantic noise and contextual misalignment. For instance, retrieving audio for a violin performance might accidentally recall cello or background applause features that, while semantically related, conflict with the specific visual cues or the user's question. To address this, we propose the context-aware adaptive purification~(CAP) mechanism. Algorithm~\ref{alg:complex_fusion} illustrates the purification process for the retrieved candidate features (e.g., visual) using the available modality (e.g., audio) as guidance. CAP functions as a dynamic semantic filter that selectively suppresses incongruent noise while substituting it with cues from common representations that are highly relevant to the question. The process is decomposed into three rigorous phases: consistency-based noise profiling, text-guided semantic acquisition, and selective feature injection.
\begin{figure*}[!t]
  \centering
  \includegraphics[width=0.95\linewidth]{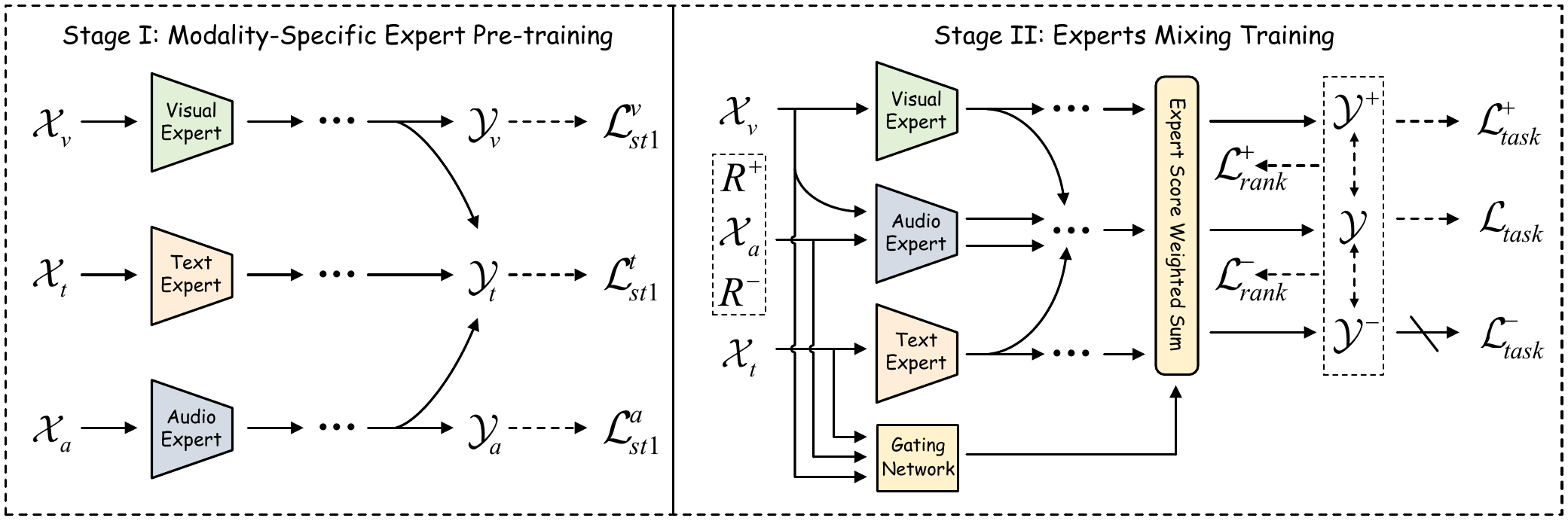}
  \caption{Two-stage training strategy sequentially performs expert pre-training and expert mixing optimization.}
  \label{fig:sub}
   \vspace{-0.8em}
\end{figure*}

\subsubsection{Consistency-based Noise Profiling}

The primary criterion for identifying noise is semantic dissonance with the available modality. Let ${H}_{avl} \in \mathbb{R}^{L \times D}$ denote the input modality representations of the available modality (e.g., visual or audio). We first abstract a global context anchor $\mathbf{g}_{avl}$ via global average pooling to capture the holistic semantic tone of the scene:
\begin{equation}
    \small
    \begin{aligned}
        \mathbf{g}_{avl} = \frac{1}{L} \sum_{t=1}^{L} {H}_{avl}[t]\\
    \end{aligned}
\end{equation}
for each retrieved token $\mathbf{r}_i \in \mathcal{R}$, we compute a dissonance score $\delta_i$, which quantifies the semantic deviation of the candidate from the current global context. To enable robust comparison in a latent manifold, we employ a learnable projection:
\begin{equation}
    \small
    \begin{aligned}
        \delta_i &= 1 - \text{sim}({H}_{miss} \cdot \mathbf{W}_{proj}, \mathbf{g}_{avl}) \\
        &= 1 - \frac{({H}_{miss} \cdot \mathbf{W}_{proj})^\top \mathbf{g}_{avl}}{\|{H}_{miss} \cdot\mathbf{W}_{proj}\| \|\mathbf{g}_{avl}\|}\\
    \end{aligned}
\end{equation}
\begin{equation}
    \small
    \begin{aligned} 
        {H}_{miss} &=  \frac{1}{n} \sum_{i=1}^{n}\mathcal{E}_{miss}(\Phi_{miss}(\mathbf{r}_i)) \\
    \end{aligned}
\end{equation}
where $\mathcal{E}_{miss}$ denotes the specific expert corresponding to the current missing modality, and $\Phi_{miss}$ represents its corresponding semantic feature encoder. Tokens with high $\delta$ values indicate retrieved information that contradicts the available evidence (e.g., retrieving ``barking'' sound for a visual ``cat''). We employ a negative selection strategy to identify the set of noise indices $\Omega_{noise}$ corresponding to the top-$k_{purge}$ discordant tokens:
\begin{equation}
    \small
    \begin{aligned}
        \Omega_{noise} = \text{Topk}_{indices}(\delta, k_{purge})\\
    \end{aligned}
\end{equation}
we further construct a binary noise mask $\mathcal{M}_{noise} \in \{0, 1\}^{L}$, where the $k$-th entry is set to 1 if $k \in \Omega_{noise}$, and 0 otherwise.

\subsubsection{Text-Guided Semantics Acquisition}

The first phase screens out erroneous content, while the second phase focuses on identifying useful information. Here, the textual question serves as a high-level semantic instruction, guiding the model to attend to specific attributes within the common knowledge $H_{com}$. We design a guidance block leveraging multi-head cross-attention~(MCA) and self-attention~(SA). Here, the retrieved candidates $\mathcal{R}$ act as the query source to seek alignment with the question intent. Let $H_{t}$ be the question embedding sequence. We compute the guidance attention map $\mathcal{A}_{self}$ and $\mathcal{A}_{cross}$ to highlight salient features:
\begin{equation}
    \small
    \begin{aligned}
        H_{guided} &= \text{ GuidanceBlock}(H_{com},H_{t}) \\
                H_{com} &= \mathcal{E}_{miss}(F_{avl}) \\
        \mathcal{A}_{self} &=  \text{SA}(Q,K,V:H_{com})\\
        \mathcal{A}_{cross} &=  \text{MCA}(Q:H_{com},K,V:H_t)\\
    \end{aligned}
\end{equation}
where $F_{avl}$ denotes the features obtained by passing the retrieved samples through the feature encoder. Simultaneously, we compute a saliency score $\sigma$ for each token by aggregating the attention weights across heads and dimensions. This score reflects the informational density of each retrieved token relative to the question. We then identify the most valuable semantic indices $\Omega_{salient}$:
\begin{equation}
    \small
    \begin{aligned}
        \Omega_{salient} = \text{Topk}_{indices}(\boldsymbol{\sigma}, k_{purge})\\
    \end{aligned}
\end{equation}
similarly, we constructed a binary mask $\mathcal{M}_{salient} \in \{0, 1\}$ from the obtained indices.

\begin{algorithm}[t!]
\small
\caption{Context-aware Adaptive Purification}
\label{alg:complex_fusion}
\textbf{Input:} audio features $F_a \in \mathbb{R}^{B \times L_a \times D}$, text features $F_t \in \mathbb{R}^{B \times L_t \times D}$, and retrieved visual features $F_v \in \mathbb{R}^{B \times L_v \times D}$. \\
\textbf{Parameter:} Purification budget $k$ (number of tokens). \\
\textbf{Output:} The Purified Visual Representations $H_{v}^{pur}$.

\begin{algorithmic}[1]
    \State Obtain representations via modality-specific experts:
    \State $H_a \leftarrow \mathcal{E}_a(F_a), \quad H_t \leftarrow \mathcal{E}_t(F_t), \quad H_v \leftarrow \mathcal{E}_v(F_v)$
    
    \State \textbf{Phase 1: Consistency-based Noise Identification}
    \State Compute global context vector $\mathbf{g}_a \in \mathbb{R}^{B \times 1 \times D}$:
    \State $\mathbf{g}_a[b] \leftarrow \frac{1}{L_a} \sum_{j=1}^{L_a} H_a[b, j, :]$
    \State Initialize similarity matrix $\delta  \in \mathbb{R}^{B \times L_v}$
    \For{$i = 1$ \textbf{to} $L_v$}
        \State $\delta[:, i] \leftarrow \text{CosineSimilarity}(\mathbf{g}_a, H_v[:, i, :])$
    \EndFor
    \State Identify noise indices set $\Omega_{noise} \subset \{1, \dots, L_v\}$ via negative selection:
    \State $\Omega_{noise} \leftarrow \text{TopK Indices}(1-\delta, k)$ \Comment{Find $k$ least correlated visual frames}
    \State Construct binary noise mask $\mathcal{M}_{noise} \in \{0, 1\}^{B \times L_v}$ based on $\Omega_{noise}$

    \State \textbf{Phase 2: Text-Guided Semantics Acquisition}
    \State Project common queries: $H_{com} \leftarrow \mathcal{E}_v(F_{a})$
    \State \emph{Guidance Mechanism (Cross-Modal \& Self-Attention):}
    \State $H_{guided}, \mathcal{A}_{cross}, \mathcal{A}_{self} \leftarrow \text{GuidanceBlock}(H_{com}, H_t)$
    \State Aggregate attention weights to estimate semantic importance:
    \State $\mathcal{W}_{total} \leftarrow \sum_{dim=-1}(\mathcal{A}_{cross}) + \sum_{dim=-1}(\mathcal{A}_{self})$
    \State Select salient semantic indices $\Omega_{salient} \subset \{1, \dots, L_c\}$:
    \State $\Omega_{salient} \leftarrow \text{TopK Indices}(\mathcal{W}_{total}, k)$ \Comment{Select $k$ most informative tokens}

    \State \textbf{Phase 3: Selective Feature Purification (Injection)}
    \For{$b = 1$ \textbf{to} $B$}
        \State Retrieve instance-specific indices: 
        \State $\mathcal{I}_N \leftarrow \Omega_{noise}[b]$, \quad $\mathcal{I}_S \leftarrow \Omega_{salient}[b]$
        \State \emph{Semantic Injection Operation:}
        \For{$j = 1$ \textbf{to} $k$}
            \State $idx_{target}, idx_{source} \leftarrow \mathcal{I}_N[j], \mathcal{I}_S[j]$
            \State $H_v[b, idx_{target}, :] \leftarrow H_{guided}[b, idx_{source}, :]$ \Comment{Overwrite noise with semantics}
        \EndFor
    \EndFor

    \State \Return $H_{v}^{pur} \leftarrow H_{v}$
\end{algorithmic}
\end{algorithm}

\subsubsection{Selective Feature Purification}

In the final phase, we perform a surgical feature replacement operation. The goal is to overwrite the identified noisy regions ($\Omega_{noise}$) with the high-quality semantic cues ($\Omega_{salient}$) extracted in second phase, while preserving the retrieval content that was deemed consistent in first phase. We construct the purified representations ${H}_{miss}^{p}$ as follows:
\begin{equation}
    \small
    \begin{aligned}
        {H}_{miss}^{pur}[i] = 
        \begin{cases} 
        H_{guided}[j], & \text{if } i \in \Omega_{noise}  \text{ and } \\ &\text{ corresponds to }  \\ &j\text{-th} \text{ salient token} \\
        H_{miss}[i], & \text{otherwise}
        \end{cases}\\
    \end{aligned}
\end{equation}
formally, this can be expressed as a masked injection operation:
\begin{equation}
    \small
    \begin{aligned}
        &{H}_{miss}^{pur} = (\mathbf{1} - \mathcal{M}_{noise}) \odot {H}_{miss} \\ &+ \mathcal{M}_{noise} \odot \text{Gather}({H}_{guided}, \Omega_{salient})\\
    \end{aligned}
\end{equation}
where $\odot$ denotes element-wise multiplication. This mechanism ensures that the final representation ${H}_{miss}^{pur}$ maintains high semantic fidelity to the real-world distribution (from $\mathcal{R}$), while simultaneously using audio-visual common knowledge to purify noise semantics that have low relevance to the current question. This purified sequence is then passed to the subsequent module for fusion.

\subsection{Two-Stage Experts Training}

To explicitly model the reliability of different information sources~(original vs. recovered), we adopt a mixture of experts architecture trained via a decoupled two-stage curriculum. The two-stage training procedure is illustrated in Figure~\ref{fig:sub}.

\noindent\textbf{Stage I: Modality-Specific Expert Pre-training}. We establish three independent experts: visual expert $\mathcal{E}_v$, audio expert $\mathcal{E}_a$, and text expert $\mathcal{E}_t$. Each expert is a transformer encoder trained to solve the AVQA task. The objective is to minimize the expert pre-training task loss:
\begin{equation}
    \small
\scalebox{0.87}{ $
    \begin{aligned}
        \mathcal{L}_{st1} = \sum_{m \in \{v, a, \{v,t,a\}\}} \mathbb{E}_{(X, y) \sim \mathcal{D}} \left[ -\log P(y | \mathcal{E}_m(F_m)) \right]\\
    \end{aligned}
$ }
\end{equation}
this ensures that the visual and audio experts extract discriminative representations ${H}_m = \mathcal{E}_m({F}_m)$ without relying on cross-modal shortcuts.

\noindent\textbf{Stage II: Expert Mixing with Dynamic Gating Training}. In the second stage, we freeze the experts and train a gating network~(Router) $\mathcal{G}$. The router dynamically assigns importance weights based on the input context. The expert weights $\boldsymbol{\alpha} \in \mathbb{R}^{3}$ are computed as:
\begin{equation}
    \small
    \begin{aligned}
        \boldsymbol{\alpha}_{m^{\prime}}&=\frac{exp({g}_{m^{\prime}})}{\sum_{m \in \{v, a, t\}}exp({g}_{m})}\\
        g_m &= \mathcal{G}(H_m,\phi)
    \end{aligned}
\end{equation}
where $\phi$ denotes the learnable parameters of the gating network $\mathcal{G}(·,\phi)$. The final joint representation is a weighted sum:
\begin{equation}
    \small
    \begin{aligned}
        \mathbf{Z}_{joint} &= \alpha_{a} H_{a} + \alpha_{t} H_t + \alpha_{v} H_{v}\\
        \mathcal{Y} &= Dec(\mathbf{Z}_{joint})\\
    \end{aligned}
\end{equation}

\subsection{Optimization and Ranking Loss}

The complete framework is optimized using a compound loss function. In addition to the standard cross-entropy loss $\mathcal{L}_{task}$, we introduce a semantic ranking loss $\mathcal{L}_{rank}$ to enforce the principle that features recovered from positive samples should be semantically superior to those from negative samples, yet inferior to the ground truth. Let $\mathcal{X}_{gt}$ denote the sample from the ground truth missing modality, and let $\mathcal{R}^+$ and $\mathcal{R}^-$ represent the retrieved positive and negative samples~(corresponding to the indices with the lowest similarity scores $S$), respectively. We impose the following constraint:

\begin{equation}
    \small
    \begin{aligned}
        \mathcal{L}_{rank}^{+} &= \max(0, \mathcal{L}_{task}(\mathcal{Y}|\mathcal{X}_{gt},y)  \\ &- \mathcal{L}_{task}(\mathcal{Y}|\mathcal{R}^+, y))\\
    \end{aligned}
\end{equation}
\begin{equation}
    \small
    \begin{aligned}
        \mathcal{L}_{rank}^{-} &= \max(0, \mathcal{L}_{task}(\mathcal{Y}|\mathcal{R}^+,y)  \\ &- \mathcal{L}_{task}(\mathcal{Y}|\mathcal{R}^-, y))\\
    \end{aligned}
\end{equation}
the total objective for expert mixing training is:  
\begin{equation}
    \small
    \begin{aligned}
        \mathcal{L}_{total} = \mathcal{L}_{task}(\mathcal{Y}, y) + \lambda (\mathcal{L}_{rank}^{+}+\mathcal{L}_{rank}^{-})\\
    \end{aligned}
\end{equation}
this optimization ensures that the retrieved and purified features lie in a valid semantic manifold that aids the QA reasoning process.

\begin{table*}[t]\centering
\renewcommand{\arraystretch}{0.65}
\setlength{\tabcolsep}{1.75mm}{
\begin{tabular}{l|c|crc|cccccc|c}
\toprule
\multirow{2}{*}{\textbf{Method}} & \multirow{2}{*}{\textbf{Venue}} & \multicolumn{3}{c|}{\textbf{Modalities}} & \multicolumn{6}{c|}{\textbf{Music-AVQA} (Audio-Visual)}  & \textbf{AVQA} \\
& & A & V & Q & Exist & Localis & Count & Comp & Temp & Avg. & Avg. \\
\midrule
\rowcolor{gray!30}\multicolumn{12}{l}{\textit{\small{AVQA Specialized Models}}} \\
PSTP-Net & MM'23 & $\bullet$ & $\bullet$ & $\bullet$ &76.18&73.23&71.80&71.79&69.00&72.57&90.20 \\
TSPM & MM'24 & $\bullet$ & $\bullet$ & $\bullet$ &82.19 & 71.85 & 76.21 & 65.76 & 71.17 & 73.51 & 90.80 \\
SHMamba & TASLP'25 & $\bullet$ & $\bullet$ & $\bullet$ & 82.89 & 67.93 & 72.65 & 61.31 & 68.37 & 70.64 & 90.80 \\
QA-TIGER & CVPR'25 & $\bullet$ & $\bullet$ & $\bullet$ & 83.10 & 72.50 & 78.58 & 63.94 & 69.59 & 73.74 & -- \\
AV-Master & arXiv'25 & $\bullet$ & $\bullet$ & $\bullet$ & 83.60 & 72.39 & 79.13 & 64.21 & 70.80 & 74.22 &91.40 \\
\rowcolor{gray!30}\multicolumn{12}{l}{\textit{\small{Incomplete Modality Learning Models}}} \\
\multirow{4}{*}{Missing-AVQA} & \multirow{4}{*}{ECCV'24} & $\bullet$ & $\circ$ & $\bullet$ & 77.94 & 58.48 & 67.43 & 65.21 & 58.88 & 65.99 & 46.65\\
 &  & $\circ$ & $\bullet$ & $\bullet$ &  78.74 & 58.15 & 70.59& 62.67 & 60.46 & 66.44 & 70.28\\ 
  & & $\bullet$ & $\bullet$ & $\bullet$ & 82.94 & 68.70 & 75.13 & 61.30 & 69.87 & 71.27 & 89.96\\
   \cline{3-12}
 &  & \multicolumn{3}{c|}{Average} & 79.87 & 61.77 & 71.05 & 63.06 & 63.07 & 67.90 & 68.96\\
 \hline
\multirow{4}{*}{MoMKE} & \multirow{4}{*}{MM'24} & $\bullet$ & $\circ$ & $\bullet$ & 79.90 & 59.61 & 65.16 & 60.32 & 58.93 & 64.82 & 59.78\\
& & $\circ$ & $\bullet$ & $\bullet$ & 78.74 & 58.15 & 70.59 & 62.67 & 61.18 & 66.44 & 70.12\\
 & & $\bullet$ & $\bullet$ & $\bullet$ & 82.24 & 67.03 & 74.96 & 63.58 & 70.16 & 71.34 & 90.26\\
  \cline{3-12}
 &  & \multicolumn{3}{c|}{Average} & 80.29 & 61.60 & 70.24 & 62.19 & 63.42 & 67.53 & 73.39\\
 \hline
\multirow{4}{*}{SimMLM} & \multirow{4}{*}{ICCV'25} & $\bullet$ & $\circ$ & $\bullet$ & 79.35 & 63.59 & 65.77 & 54.46 & 68.49 & 65.64 & 59.95\\
& & $\circ$ & $\bullet$ & $\bullet$ & 81.07 & 60.17 & 69.25 & 63.03 & 60.46 & 66.94 & 70.88\\
 & & $\bullet$ & $\bullet$ & $\bullet$ & 82.78 & 66.92 & 75.81 & 64.02 & 69.24 & 71.57 & 90.32\\
  \cline{3-12}
 &  & \multicolumn{3}{c|}{Average} & 81.07 & 63.56 & 70.28 & 60.50 & 66.06 & 68.05 & 73.72\\
 \hline
\multirow{4}{*}{IMOL} & \multirow{4}{*}{ACL'25} & $\bullet$ & $\circ$ & $\bullet$ & 81.10 & 61.33 & 70.83 & 61.16 & 61.95 & 67.11 & 61.32\\
& & $\circ$ & $\bullet$ & $\bullet$ & 81.72 & 65.87 & 69.80 & 61.98 & 68.25 & 69.21 & 72.38 \\
 & & $\bullet$ & $\bullet$ & $\bullet$ & 83.49 & 70.35 & 74.68 & 63.21 & 69.10 & 71.86 & 90.28\\
  \cline{3-12}
 &  & \multicolumn{3}{c|}{Average} & 82.10 & 66.18 & 71.77 & 62.12 & 66.43 & 69.39 & 74.66\\
 \hline
\multirow{4}{*}{\textbf{R$^2$ScP~(ours)}} & \multirow{4}{*}{--} & $\bullet$ & $\circ$ & $\bullet$ &81.07&65.54&71.32&61.95&67.51&69.37&63.25\\
 &  & $\circ$ & $\bullet$ & $\bullet$ &82.49&68.93&74.05&63.49&69.34&72.06&75.12\\
 & & $\bullet$ & $\bullet$ & $\bullet$ &83.59&70.57&75.95&64.08&70.75&73.19&90.64\\
 \cline{3-12}
 &  & \multicolumn{3}{c|}{Average} 
 & \cellcolor{blue!10}\textbf{82.38}
 & \cellcolor{blue!10}\textbf{68.35}
 & \cellcolor{blue!10}\textbf{73.77}
 & \cellcolor{blue!10}\textbf{63.17}
 & \cellcolor{blue!10}\textbf{69.61}
 & \cellcolor{blue!10}\textbf{71.54}
 & \cellcolor{blue!10}\textbf{76.35}
\\
\bottomrule
\end{tabular}}
\caption{Comparison of different methods on Music-AVQA and AVQA dataset under various missing modality settings~($\circ$ indicates a missing modality). A: audio, V: visual, Q: question. Exist, Localis, etc. represent the accuracy in the subtasks of the Music-AVQA dataset.}
\label{MUSIC-AVQA}
\vspace{-0.3cm}
\end{table*}

\begin{table}[t]
  \centering
  \renewcommand{\arraystretch}{0.7}
  \scalebox{1.0}{
    \begin{tabular}{cc|cc|c|c}
    \toprule
     \multicolumn{2}{c}{Modalities} & \multicolumn{2}{c|}{Modules} & \multirow{2}[4]{*}{\makecell[c]{MUSIC\\AVQA}} & \multirow{2}[4]{*}{AVQA} \\
\cmidrule{1-4}        A     & V     & CMR    & CAP    &  \\
    \midrule
     $\bullet$ & $\circ$ &  &  &  62.43 & 57.43 \\
     $\bullet$ & $\circ$ &  & \checkmark &  64.11 & 59.64 \\
     $\bullet$ & $\circ$ & \checkmark &  &  67.21 & 61.78 \\
    \rowcolor{red!10}
     $\bullet$ & $\circ$ & \checkmark & \checkmark  &  69.37 & 63.25 \\
    \midrule
     $\circ$ & $\bullet$ &  &   & 63.54 & 68.02\\
    $\circ$ & $\bullet$ &  & \checkmark  & 65.21 & 69.14\\
     $\circ$ & $\bullet$ & \checkmark &   & 70.18 & 73.86\\
    \rowcolor{red!10}
     $\circ$ & $\bullet$ & \checkmark & \checkmark   & 72.06 & 75.12 \\
    \midrule
     $\bullet$ & $\bullet$ &  &   & 71.14  & 88.68\\
     $\bullet$ & $\bullet$ &  & \checkmark  & 72.12 & 89.43\\
     $\bullet$ & $\bullet$ & \checkmark &   & 72.42 & 90.06\\
    \rowcolor{red!10}
    $\bullet$ & $\bullet$ & \checkmark & \checkmark  & 73.19 & 90.64\\
    \bottomrule
    \end{tabular}%
    }
    \caption{Ablation on the different components of R$^2$ScP.}
    \vspace{-0.5cm}
  \label{tab:ablation1}%
\end{table}%

\section{Experiments}

In this section, we conduct experiments to evaluate the effectiveness of the proposed model. We also compare our method with a range of related methods, including MoE-based approaches~(IMOL~\cite{zeng2025imol}, SimMLM~\cite{li2025simmlm}, MoMKE~\cite{xu2024leveraging}) and other architectures~(Missing-AVQA~\cite{park2024learning}).

\subsection{Performance Results}

We evaluate the effectiveness of the proposed R$^2$ScP framework by comparing it with state-of-the-art methods on two benchmarks: Music-AVQA~\cite{li2022learning}, AVQA~\cite{yang2022avqa}. As presented in Table~\ref{MUSIC-AVQA}, we report the performance comparison on the Music-AVQA and AVQA datasets under the missing modality setting. The results demonstrate that R$^2$ScP consistently outperforms existing state-of-the-art counterparts across both datasets. Specifically, across various modality settings, R$^2$ScP achieves a new state-of-the-art average accuracy of 71.54\% on Music-AVQA and 76.35\% on AVQA. Notably, the performance gain is more pronounced on the AVQA dataset, which encompasses a diverse range of open-domain daily events. We attribute this to the fact that generative methods often struggle to synthesize realistic features for complex, unconstrained scenes. In contrast, our R$^2$ScP framework effectively retrieves and filters real-world semantic cues, thereby preserving more distinct modality-specific details for robust reasoning. Furthermore, compared to several AVQA specialized models, our method demonstrates competitive performance even under the full modality setting, despite our primary focus on learning with missing modalities.

\begin{figure}[t]
  \centering
  \includegraphics[width=\linewidth]{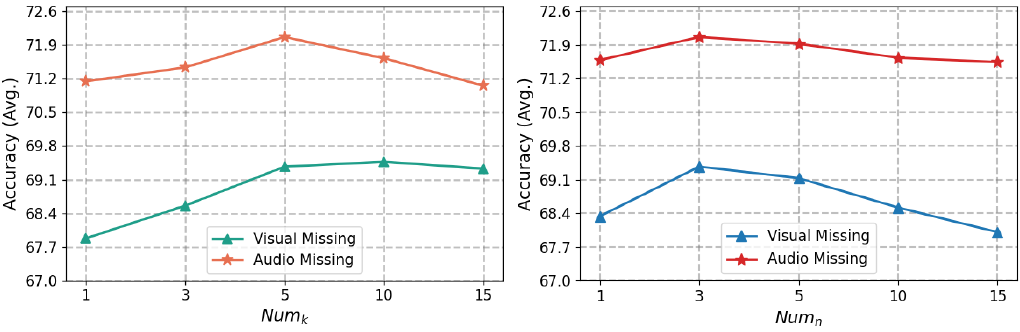}
  \caption{Impact of purification budget $k$ and number of retrieved samples $n$.}
  \label{fig:3}
  \vspace{-0.2cm}
\end{figure}

\begin{table}[t]
\centering
\renewcommand{\arraystretch}{0.8}
\begin{tabular}{lc}
\hline
Method & Avg. \\ \hline
R$^2$ScP~(ours) & \textbf{70.72} \\
w/o modality-specific expert pretraining & 68.98 \\
w/o expert mixing training & 64.21 \\
w/o ranking loss & 69.62 \\ 
\hline
\end{tabular}
\caption{Effectiveness of the  two-stage training strategy. Results are the average performance on Music-AVQA for the visual and audio missing settings.}
\vspace{-0.5cm}
\label{tab:ablation2}
\end{table}

\subsection{Ablation Study}

\begin{figure*}[t]
  \centering
  \includegraphics[width=\linewidth]{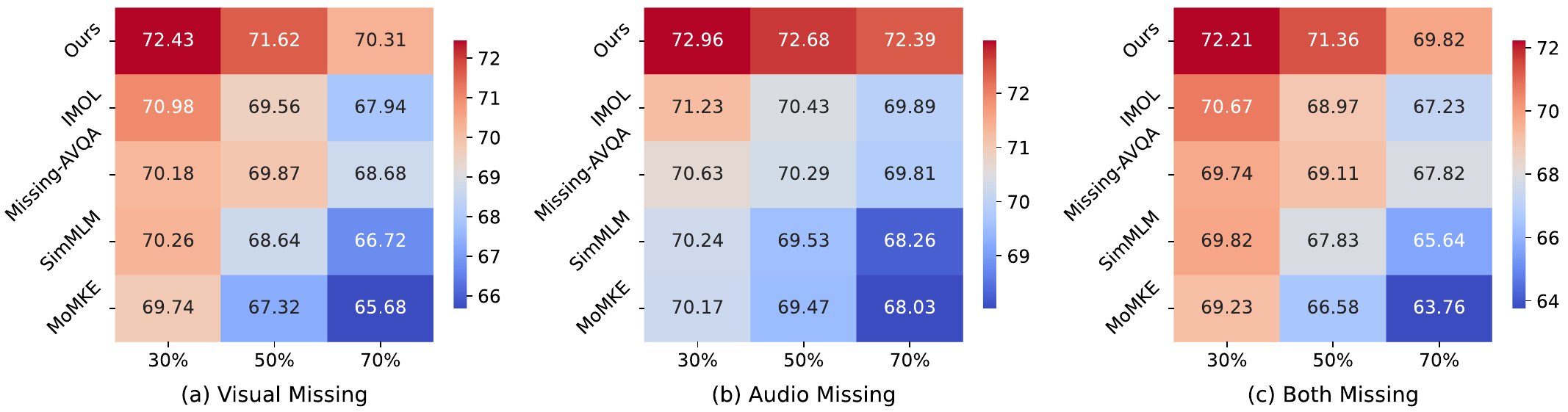}
  \caption{Generalization analysis on the Music-AVQA dataset across various missing rates}
  \label{fig:4}
  \vspace{-0.3em}
\end{figure*}

To better understand the individual contributions of each proposed component, we conduct a comprehensive ablation study of the proposed R$^2$ScP.


\noindent \textbf{Effectiveness of retrieval and purification modules}. We first investigate the necessity of the Cross-Modal Retrieval (CMR) and Context-Aware Adaptive Purification (CAP) mechanisms in Table~\ref{tab:ablation1}. The baseline model, which relies solely on available modalities without retrieval, suffers significant performance degradation in missing modality scenarios. Notably, employing CAP alone enhances the decoding process by leveraging common semantics to refine available features. Furthermore, the CMR module yields substantial gains by retrieving external cues from a unified semantic space to compensate for information loss. Combining both achieves the highest accuracy. This demonstrates that while retrieval offers essential raw evidence, CAP is indispensable for filtering semantic noise to synthesize high-fidelity representations. 

\begin{table}[t]
\centering
\begin{tabular}{c|c|c}
\hline
Corpus & Music-AVQA & AVQA \\ \hline
N/A & 64.66 & 64.39 \\
AVQA & 66.51 & 69.19 \\
VGGSound & 66.92 & 69.64 \\
Music-AVQA & 70.72 & 65.72 \\
\hline
\end{tabular}
\caption{Impact of different retrieval corpora on the performance of Music-AVQA and AVQA datasets. Results
are the average performance for the
visual and audio missing settings.}
\label{tab:Corpus}
\end{table}


\noindent \textbf{Ablation on training strategies.} We validate our optimization protocol in Table~\ref{tab:ablation2}. The most significant drop from 70.72\% to 64.21\% occurs without expert mixing (Stage II), underscoring the vital role of the MoE gating network. Bypassing independent pre-training (Stage I) also degrades performance to 68.98\%, indicating that experts require strong unimodal foundations before learning cross-modal dependencies. Finally, the absence of the ranking loss $\mathcal{L}_{rank}$ reduces accuracy to 69.62\%. This confirms that $\mathcal{L}_{rank}$ is essential for enforcing a structured semantic manifold to aid the purification process, as standard task losses alone are insufficient for ensuring semantic quality of retrieved features.

\subsection{Analysis and discussion}

To provide a deeper understanding of the internal mechanisms and robustness of R$^2$ScP, we conduct a detailed analysis concerning hyperparameter sensitivity and performance generalization under varying degrees of data incompleteness.


\noindent\textbf{Impact of purification budget and retrieval count}.
Figure~\ref{fig:3} illustrates the sensitivity of retrieval count $Num_{n}$ and purification budget $Num_{k}$ on Music-AVQA. Regarding $Num_{n}$, accuracy peaks at 3 and subsequently declines because excessive candidates introduce semantic noise that confounds expert reasoning. For $Num_{k}$, we observe an inverted U-shaped trend peaking at 5. A low budget fails to inject sufficient common semantics for noise rectification, while an overly aggressive budget risks overwriting valid context-aligned information in the original representation.

\noindent\textbf{Impact of varying missing ratios}. Real-world scenarios often involve varying degrees of severe data loss. To evaluate the robustness of R$^2$ScP, we test the model performance across three distinct missing rates: 30\%, 50\%, and 70\%. As illustrated in Figure~\ref{fig:4}, R$^2$ScP consistently outperforms all competing methods across all missing rates and modality scenarios. Notably, the performance gap widens as the missing rate increases. Existing baseline methods like Missing-AVQA and SimMLM exhibit a steeper performance degradation slope as the missing rate climbs from 30\% to 70\%. This validates our hypothesis that generative imputation methods struggle when the available context is scarce, leading to hallucination bottlenecks. In contrast, by retrieving real-world knowledge from an external semantic space, R$^2$ScP mitigates the dependency on the immediate context, demonstrating superior stability even when the majority of the modality data is absent.


\begin{figure}[!t]
  \centering
  \includegraphics[width=\linewidth]{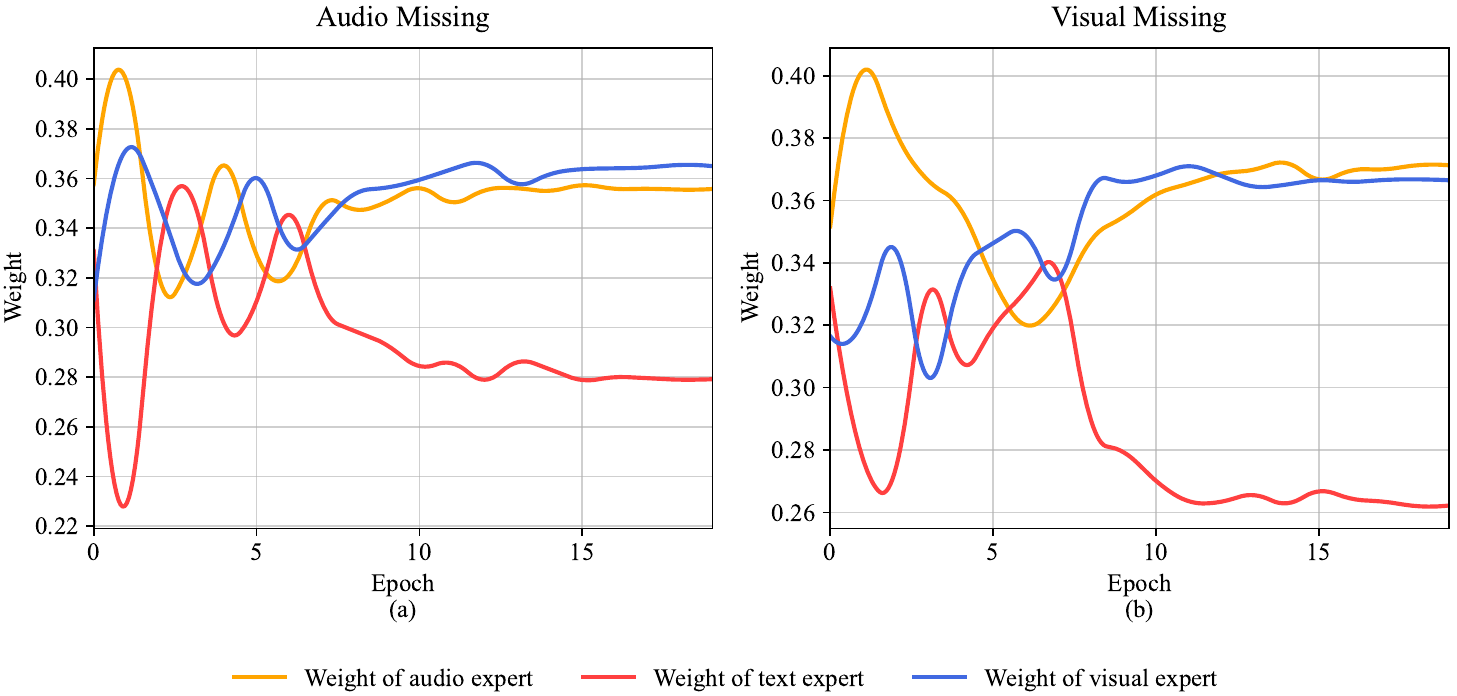}
  \caption{Trends in expert loads during experts mixing training on Music-AVQA in the severely incomplete conditions.}
  \label{fig:expert Visualization}
\end{figure}

\begin{figure*}[t]
  \centering
  \includegraphics[width=\linewidth]{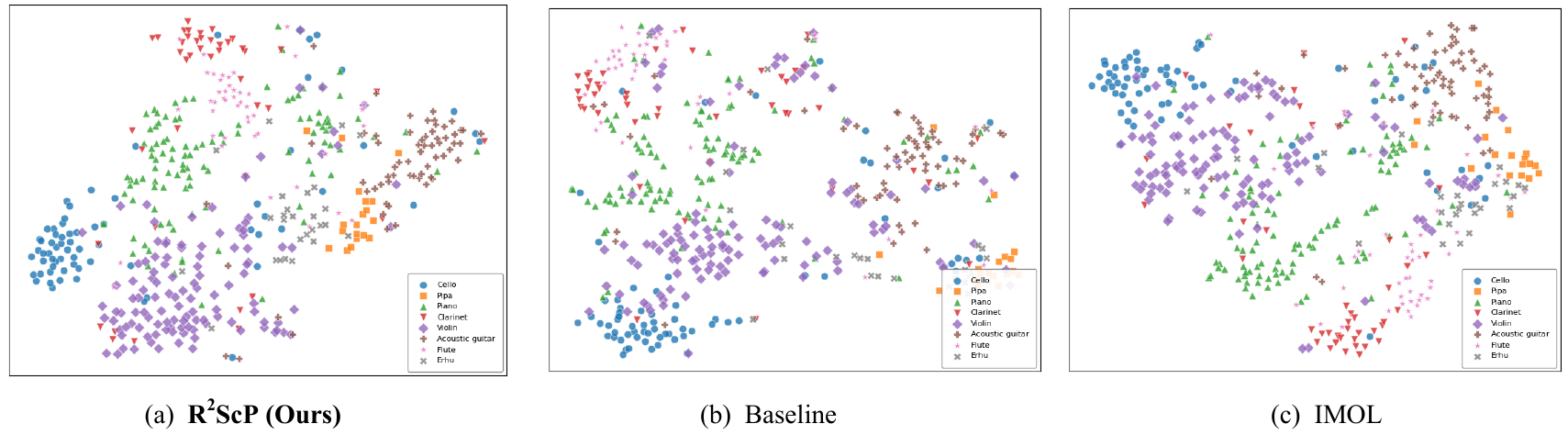}
  \caption{t-SNE visualization of our model and other methods on the Music-AVQA dataset.}
  \label{fig:5}
\end{figure*}

\noindent \textbf{Impact of retrieval corpus.} We analyze the influence of corpus domain and scale in Table~\ref{tab:Corpus}. Incorporating any external corpus consistently surpasses the baseline, validating the efficacy of knowledge compensation. Crucially, domain consistency proves vital as the model achieves an optimal accuracy of 70.72\% using the in-domain Music-AVQA corpus. Performance declines when shifting to natural scene corpora like AVQA (66.51\%) or VGGSound (66.92\%). However, VGGSound outperforms AVQA within the same domain~(natural scenes). This confirms that larger data scales increase the likelihood of retrieving high-quality candidates to enhance performance.

\noindent \textbf{Analysis of expert loads}. To gain insights into the mechanism of the mixture-of-experts framework under incomplete modality conditions, we further visualize the expert loads on the test set during the expert mixing training process, as illustrated in Figure~\ref{fig:expert Visualization}. These loads correspond to the importance weights dynamically assigned by the Soft Router to each modality-specific expert. We observe that as the training progresses, the model gradually converges, and the expert loads tend to stabilize after initial fluctuations. Notably, the features recovered via our retrieval mechanism are assigned importance weights comparable to those of the currently available high-level semantic modalities (e.g., Visual or Audio). This demonstrates that the R$^2$ScP framework effectively treats the retrieved knowledge as a reliable semantic source, successfully bridging the information gap caused by missing modalities.

\subsection{Visualization}

Figure~\ref{fig:5} illustrates the t-SNE~\cite{maaten2008visualizing} visualization of embedding distributions on the Music-AVQA test set under visual modality missing scenarios. Compared to the baseline (w/o CMR and CAP) and IMOL, the feature points corresponding to R$^2$ScP exhibit significantly tighter clustering and more distinct separability. This strongly evidences the efficacy of R$^2$ScP in handling modality missingness. Specifically, the high density of the clustered points indicates the model's accuracy in recognizing similar samples. Furthermore, the clear boundaries between different answer categories mitigate misclassification. Concurrently, the uniform distribution suggests that R$^2$ScP effectively balances inter-class relationships, resulting in more stable and reliable model outputs. These observations validate that R$^2$ScP maintains robust comprehension capabilities despite the challenges of missing data, highlighting its superiority in incomplete modality learning.

\section{Conclusion}

In this paper, we present R$^2$ScP, a framework that addresses missing modalities in AVQA by shifting from generative imputation to retrieval-based recovery. We introduce the Context-aware Adaptive Purification (CAP), which dynamically filters semantic noise from retrieved data by enforcing consistency with the available modalities for high-fidelity reconstruction.  Extensive experiments conducted on multiple benchmarks confirm that R$^2$ScP outperforms state-of-the-art methods, demonstrating superior accuracy and robustness. Future work will explore larger-scale retrieval databases to further enhance generalization capabilities.

\section{Acknowledgments}

This work was supported by National Natural Science Foundation of China (Grant No. 62306061 and 62576076), CCF-Tencent Rhino-Bird Open Research Fund. The computational resources are supported by SongShan Lake HPC Center (SSL-HPC) in Great Bay University.

\section{Limitations} This study suffers limitations that may impact the
performance of our proposed framework. Although retrieval-based missing modality recovery strategies have demonstrated effectiveness in audio-visual question answering, the model's inference accuracy remains sensitive to the quality of the retrieved samples. Furthermore, our current work addresses missing modalities exclusively during the inference phase. However, in practical applications, missing modalities may also occur during the learning process~(training). Consequently, future work will aim to address this by developing AVQA models that are robust to missing modalities during the training stage as well.


\bibliography{main}

\appendix



\end{document}